\documentclass{article}

\usepackage{arxiv}

\usepackage[utf8]{inputenc}
\usepackage[T1]{fontenc}   
\usepackage{hyperref}      
\usepackage{url}           
\usepackage{booktabs}      
\usepackage{amsfonts}
\usepackage{amsmath}      
\usepackage{nicefrac}      
\usepackage{microtype}     
\usepackage{cleveref}      
\usepackage{graphicx}
\usepackage[square, numbers]{natbib}
\usepackage{doi}
\usepackage{caption}
\usepackage{subcaption}
\usepackage{bm}
\usepackage{algorithm}
\usepackage{algpseudocode}
\usepackage{setspace}
\usepackage[inkscapepath=svgsubdir]{svg}
\usepackage{tipa}

\usepackage{tikz}
\usetikzlibrary{calc,patterns,decorations.pathmorphing,decorations.markings,positioning,decorations.shapes,decorations}
\usetikzlibrary{arrows,arrows.meta,bending,scopes}
\usetikzlibrary{quotes,angles}

\pgfdeclaredecoration{damping}{initial}
{
\state{initial}[width=(\pgfdecoratedpathlength)/4,next state=dbottom]
{
\pgfpathlineto{\pgfpoint{(\pgfdecoratedpathlength)/4}{0}}
}
\state{dbottom}[width=(\pgfdecoratedpathlength)/4,next state=final]{
  \pgfpathlineto{\pgfpoint{0pt}{(\pgfdecoratedpathlength)/1.618/4}}
  \pgfpathlineto{\pgfpoint{(\pgfdecoratedpathlength)/2}{(\pgfdecoratedpathlength)/1.618/4}}
  \pgfpathlineto{\pgfpoint{0pt}{(\pgfdecoratedpathlength)/1.618/4}}
  \pgfpathlineto{\pgfpoint{0pt}{-(\pgfdecoratedpathlength)/1.618/4}}
  \pgfpathlineto{\pgfpoint{(\pgfdecoratedpathlength)/2}{-(\pgfdecoratedpathlength)/1.618/4}}
  \pgfpathmoveto{\pgfpoint{(\pgfdecoratedpathlength)/4}{0pt}}
}
\state{final}
{
  \pgfpathlineto{\pgfpoint{0pt}{(\pgfdecoratedpathlength)/8}}
  \pgfpathlineto{\pgfpoint{0pt}{-(\pgfdecoratedpathlength/8}}
  \pgfpathlineto{\pgfpoint{0pt}{0pt}}
  \pgfpathlineto{\pgfpointdecoratedpathlast}
}
}

\tikzstyle{force}=[thick,-Latex]
\tikzstyle{spring}=[thick,decorate,decoration={zigzag,pre length=0.3cm,post length=0.3cm,segment length=6}]
\tikzstyle{damper2}=[thick,decoration=damping,decorate]
\tikzstyle{damper}=[thick,decoration={markings,
  mark connection node=dmp,
  mark=at position 0.25 with
  {
    \node (dmp) [thick,inner sep=0pt,transform shape,rotate=-90,minimum width=15pt,minimum height=3pt,draw=none] {};
    \draw [thick] ($(dmp.north east)+(10pt,0)$) -- (dmp.south east) -- (dmp.south west) -- ($(dmp.north west)+(10pt,0)$);
  }
}, decorate]
\tikzstyle{ground}=[fill,pattern=north east lines,draw=none,minimum width=0.75cm,minimum height=0.3cm]

\newcommand{\B}[1]{\bm{#1}}
\newcommand{\m}[1]{\begin{bmatrix} #1 \end{bmatrix}}
\renewcommand*{\d}{\mathop{}\mathrm{d}}
\newcommand{\tran}{^{\mkern-1.5mu\mathsf{T}}}

\title{PAO: A general particle swarm algorithm with exact dynamics and closed-form transition densities}
\renewcommand{\shorttitle}{PAO: A general particle swarm algorithm with exact dynamics}
\date{April, 2023}
\author{
	\href{https://orcid.org/0000-0002-3037-7584}{\includegraphics[scale=0.06]{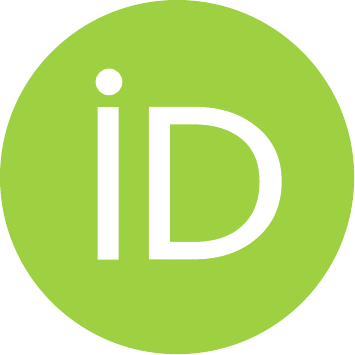}\hspace{1mm}Max D.~Champneys}
	\thanks{Corresponding author} \\
	Dynamics Research Group\\
	University of Sheffield\\
	Mappin St, Sheffield, S1 4ET\\
	\texttt{max.champneys@sheffield.ac.uk} \\
	\And
	\href{https://orcid.org/0000-0002-3433-3247}{\includegraphics[scale=0.06]{orcid.pdf}\hspace{1mm}Timothy J.~Rogers} \\
	Dynamics Research Group\\
	University of Sheffield\\
	Mappin St, Sheffield, S1 4ET\\
	\texttt{tim.rogers@sheffield.ac.uk} \\
}
\hypersetup{
	pdftitle={\shorttitle},
	pdfsubject={Particle swarm optimisation},
	pdfauthor={Max D Champneys},
	pdfkeywords={Heuristic optimisation, Particle swarm optimisation, Stochastic differntial equations},
}

\begin{document}

\maketitle
\begin{abstract}
	A great deal of research has been conducted in the consideration of meta-heuristic optimisation methods that are able to find global optima in settings that gradient based optimisers have traditionally struggled. Of these, so-called particle swarm optimisation (PSO) approaches have proven to be highly effective in a number of application areas. Given the maturity of the PSO field, it is likely that novel variants of the PSO algorithm stand to offer only marginal gains in terms of performance---there is, after all, no free lunch. Instead of only chasing performance on suites of benchmark optimisation functions, it is argued herein that research effort is better placed in the pursuit of algorithms that also have other useful properties. In this work, a highly-general, interpretable variant of the PSO algorithm ---particle attractor algorithm (PAO)--- is proposed. Furthermore, the algorithm is designed such that the transition densities (describing the motions of the particles from one generation to the next) can be computed exactly in closed form for each step. Access to closed-form transition densities has important ramifications for the closely-related field of Sequential Monte Carlo (SMC). In order to demonstrate that the useful properties do not come at the cost of performance, PAO is compared to several other state-of-the art heuristic optimisation algorithms in a benchmark comparison study.
\end{abstract}


\section{Introduction}

A great number of challenges in science, engineering and beyond can be cast as optimisation problems. Indeed, a great deal of research effort has been expended in the specification of methods that are able to provide optimal parameters for a given objective function. Of particular interest are approaches that can perform the optimisation without access to gradient information. So called \emph{heuristic methods} have attracted a great deal of research interest over many decades. Already, countless methods have been proposed, ranging from early genetic algorithms \cite{Koza1994} to more recent evolutionarily inspired approaches. Recent comprehensive reviews of the field can be found in \cite{AbdelBasset2018, Hussain2019}. Despite strong empirical performance, the no-free-lunch theorem for optimisation \cite{Joyce2018} precludes the existence  of a single `best' approach. For any two optimisation algorithms it is always possible to find an optimisation task upon which one algorithm outperforms the other (and \emph{vice versa}).

Although no algorithm can be proven to be better than any other, this does not mean that new approaches should not be sought out. Instead of placing the focus solely on benchmark performance, it is useful to consider other factors that make some heuristic methods more useful in practice. In this work, a novel methodology is proposed that incorporates both \emph{interpretable hyperparameters}, \emph{exact dynamics} and \emph{closed-form transition densities}.

One issue (amongst many \cite{Kudela2023}) with several heuristic and meta-heuristic optimisation strategies is that the performance of the approach comes to depends strongly on the choice of several hyperparameters. In many optimisation settings there exists a trade-off between so-called exploration (sampling the search space for promising minima) and exploitation (greedily searching within a minima for an optima) behaviours. Hyperparameter choice can play a key role in biasing the search to one mode of optimisation or the other. A motivation of the current work is that this trade-off is made as interpretable as possible.

A further motivation for the approach presented in this work is direct closed-form access to the transition densities of the particles. The position of any particle in the next generation can be expressed as by a Gaussian distribution that depends only on the previous value. This is desirable as it enables the optimisation process to be used as a  proposal step within a sequential Monte Carlo (SMC) scheme \cite{Chopin2020} in order to provide robust uncertainty quantification in the presence of noisy objective functions.

\subsection{Background and related work}


Optimisation algorithms inspired by nature and evolution have been a a topic of considerable research since the early twentieth century. Although many algorithms have been proposed, there has been recent concern that many of these offer at best incremental novelty and at worst a deliberate deception due to inherent bias towards optimisation benchmark functions where the optima lie at the centre of the search space \cite{Kudela2023}. Although many new algorithms are published every year, many application papers rely on a core number of highly-cited methods and their variants including ant colony optimisation \cite{Dorigo1997}, particle swarm optimisation (PSO) \cite{Kennedy1995} and differential evolution (DE) \cite{Storn1997}.


Among the most intensively researched heuristics are those belonging to the particle swarm class. First proposed by Kennedy and Eberhart in \cite{Kennedy1995}, the approach has been extensively studied and extended. Applications of particle swarm methods can be found in diverse fields from power systems \cite{DelValle2008} to image segmentation \cite{Omran2006}. Since the original proposition, a great deal of variants to the method have been proposed, including additive noise \cite{Kennedy}, quantum leaps \cite{Sun2004}, Gaussian attractors \cite{Sun2011}, ordinary differential equation attractors in \cite{Yang2014} and multiple constraints in \cite{Duan2022} as well as many others.


In this study, the PSO algorithm is treated as a stochastic differential equation (SDE). Casting the particle swarm as an SDE is not in itself a novel approach. In a series of papers Grassi et al. \cite{Grassi2020, grassi2021mean} cast several variants of the particle swarm algorithm as SDEs in order to derive convergence results from mean-field limit theory. However, the derivations in \cite{Grassi2020, grassi2021mean} result in nonlinear SDEs that cannot be solved exactly in order to recover closed-form transition densities.


\subsection{Contribution}

Although particle swarms have been separately cast as SDEs \cite{Grassi2020, grassi2021mean} and employed with additive stochastic terms \cite{Kennedy}, the authors are not presently aware of any methods that have employed both of these approaches with the specific aim of providing a particle swarm algorithm with a closed-form transition density.

In this paper, the authors propose a variant PSO approach -- the particle attractor optimisation (PAO)\footnote{Pronounced `pa\textupsilon'.} algorithm. In the proposed approach, the motions of each element of each particle are given by the solution of a scalar, stochastic differential equation with additive noise. Although the formulation bears a close resemblance to existing approaches, the linear stochastic formulation here comes with a number of distinct advantages over existing particle swarm methods:

\begin{itemize}
	\item The dynamics of the particles are computed exactly and no difference-equation approximation is required.
	\item The method is highly flexible through the choice of the attractors.
	\item The hyperparameters of the approach are interpretable as the parameters of a dynamic system (damping ratio, natural frequency etc.).
	\item The forward transition densities of the particles are available in closed form.
\end{itemize}

\section{Methodology}

In this work, a generalised version of a particle swarm optimisation algorithm is proposed that enables the particle dynamics from one iteration to the next to be represented as a linear, time-invariant SDE with additive stochasticity. As in other PSO algorithms, the motions of the particles are driven by restoring forces that act in the direction of a number of attractors in the search space. The positions of the attractors are user-defined and updated each iteration. Choices for the attraction points might include:

\begin{itemize}
	\item \emph{Global best}: The best particle location found across all particles.
	\item \emph{Local best}: The best location in the location history of a single particle.
	\item \emph{Average local best}: The local best positions averaged across all particles\footnote{Care must be taken to avoid introducing centre-bias to the optimisation.}.
	\item \emph{Average particle}: The average particle location.
	\item \emph{Weighted average particle}: The average particle location weighted by the fitness scores.
	\item Attractors based on other meta-heuristic methods i.e `rand/1/bin' from differential evolution \cite{Storn1997}.
	\item Stochastic attractors such as those in \cite{Sun2011}.
\end{itemize}

\begin{figure} 
	\centering
	\begin{tikzpicture}
		
		\draw [cyan] plot [smooth, tension=0.8] coordinates {(-4,-2) (-3,-1.8) (-1.5,-0.5) (0,0)};
				
		\node[draw, circle, fill] (particle) at (0,0) {};
		\node (label) at ($(particle)+(0,-0.4)$) {$\B{x}$};
		\draw[force] (particle) -- ++ (-1, 1) node [pos=1.25] {$\epsilon\sim\mathcal{N}(0,\sigma^2)$};
		\draw[force] (particle) -- ++ (-1.4, -0.1) node [pos=1.75] {$m\ddot{\B{x}}+c\dot{\B{x}}$};
		\draw[spring] (particle) -- ++ (3,2) node [pos=1, fill, circle,inner sep=0pt,minimum size=3pt] {};
		\draw[spring] (particle) -- ++ (3,-2) node [pos=1, fill, circle, inner sep=0pt,minimum size=3pt] {};
		\node (a1) at ($(particle) + (3.5,2)$) {$\alpha_1$};
		\node (a2) at ($(particle) + (3.5,-2)$) {$\alpha_2$};
		\draw[dashed] ($(particle) + (3,2)$) -- ($(particle) + (3,-2)$) node[midway, right, ] {$\alpha^\prime$};
		\draw[dashed] ($(particle) + (3,2)$) -- ($(particle) + (3,-2)$) node[midway, fill, circle, minimum size=3pt, inner sep=0pt ] {};

		\node at ($(particle) + (1.5,1.5)$) {$k_1$};
		\node at ($(particle) + (1.5,-1.5)$) {$k_2$};

		\draw[force] (particle) -- ++ (2,0) node [pos=1.25] {$k^\prime\B{x}^\prime$};
		\draw[red, -stealth, thick] (particle) -- ++ (1.6,0.5) node [pos=1.25] {$\Delta\B{x}$};
		
	\end{tikzpicture}
	\caption{Visualisation of the motions of a single particle in the proposed algorithm. The particle moves as if attached to linear elastic springs at each attraction point. The movement of the particle is subject to inertial and viscosity terms that resist motion. Stochasticity is introduced as a Gaussian white noise excitation term. The red arrow represents the restoring force from equivalent single spring that acts towards the overall attractor weighted by the stiffness terms.}
	\label{fig:algo}
\end{figure}
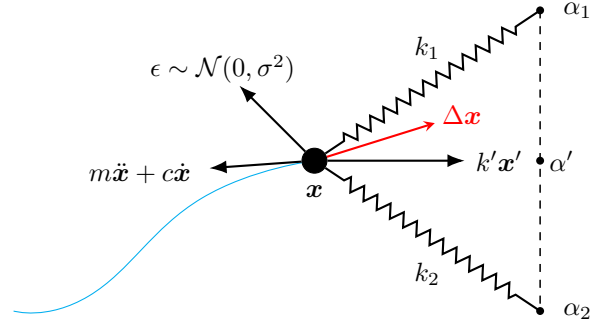

However, the formulation of the approach here does not depend on the choice of attractors. In order to ensure that the underlying SDE has an exact solution, stochasticity is introduced to the dynamics by the application of independent forcing terms that are applied to each element of each particle independently and additively. The noise process is modelled as a zero-mean Gaussian with variance given by,

\begin{equation}
	\sigma^2 = q\nu(\alpha)
\end{equation}

where $\nu$ is a user-defined function that depends only on the positions of the attractors, scaled by the hyperparameter $q_0$. In this work, $\nu(\alpha)$ is defined as the sum of the squared distances between the average and global best particle locations. The reasoning here is that the amount of noise should decrease as these attraction centres converge onto each other.
The forces acting on each particle are depicted in Figure \ref{fig:algo}.

Mathematically, the motions of the particles can be described by a second-order differential equation. In order to simplify the forthcoming notation, let, $x = x_{ij}$ be the position of the $j$th element of the $i$th particle in the search space. Similarly, let, $\alpha_r = \alpha_{rij}$ be the position of the $j$th element of the $i$th particle for the $r$th attractor. Because the dynamics of the motions of the elements of the particles are all independent from each other, they can be individually represented by the second-order scalar differential equation,

\begin{equation}
	m \ddot{x} + c \dot{x} + \sum_r k_r (x - \alpha_r) + w(t)
\end{equation}

where $m$, $c$ and $k_i$ are hyperparameters relating to the inertia, viscous drag and relative weights of the attractors $\alpha_r$ and each over-dot represents a time derivative. $w(t)$ is a continuous time white-noise process with zero mean and spectral density given by $\sigma^2$.

The equation of motion above can be trivially recast into a first-order state-space form as,

\begin{equation}
	\dot{\B{x}} = \m{\dot{x}' \\ \dot{z}} = \m{z \\ \frac{1}{m} (-cz -k'x' -\epsilon)} = \B{0}
\end{equation}

where the following coordinate transformations have been made,

\begin{equation}
	k' = \sum_r k_r
\end{equation}

\begin{equation}
	x' = x - \frac{1}{k'}\sum_r k_r \alpha_r
\end{equation}

\begin{equation}
	z = \dot{x}'
\end{equation}

To aid with the interpretability of the hyperparameters, it is convenient to express the viscous damping term in the above in terms of the damping ratio \cite{thomson1996theory},

\begin{equation}
	\zeta = \frac{c}{2\sqrt{k'm}}
\end{equation}

The above differential equation admits an Itô process representation of the form,

\begin{equation}
	\d \B{x} = F \B{x} \d t + \B{L} \d \beta
	\label{eqn:ito_form}
\end{equation}

whereby

\begin{equation}
	F = \m{0 & 1 \\ -\frac{k'}{m} & -2\sqrt{\frac{k'}{m}}\zeta}
\end{equation}

\begin{equation}
	\B{L} = \m{0 \\ 1}
\end{equation}

and $\beta$ is the formal Brownian motion process in one dimension with diffusion coefficient given by $Q=\sigma^2$ (the spectral density of the driving white noise process). If one assumes that all $m$, $c$, $k'$, $\alpha_r$ and $q$ are constant in the interval $[t_0, t]$,\footnote{This is an implicit assumption in the discretisation of the dynamics in practically every particle swarm optimisation method.} then the solution to \eqref{eqn:ito_form} is a standard result in the SDE literature \cite{Saerkkae2019}. In order to keep this paper self contained, the salient results are reproduced here. The general solution to \eqref{eqn:ito_form} is given by,

\begin{equation}
	\mathbf{x}(t) = \exp \left(F\left(t-t_0\right)\right) \B{x}\left(t_0\right)+\int_{t_0}^t \exp (F(t-\tau)) \B{L} \mathrm{d} \beta(\tau)
\end{equation}

where $\exp(\cdot)$ represents the matrix exponential operation. Denoting now,

\begin{equation}
	A = \exp \left(F \Delta t\right) = \exp \left(F\left(t-t_0\right)\right)
\end{equation}

then the time evolution of each element of each particle is given by,

\begin{equation}
	\B{x}_t = A \B{x}_{0}+\int_0^{\Delta t} A \B{L} \mathrm{d} \beta
\end{equation}

where $\B{x}_0$ is a given (deterministic) initial condition. The mean and covariance of the above can now be computed,

\begin{equation}
	\mathbb{E}[\B{x}_t] = A \B{x}_0
\end{equation}

\begin{equation}
	\mathbb{E}[\B{x}_t \tran \B{x}_t] =  \int_0^{\Delta t} \exp\left(F\left(\Delta t - \tau\right)\right) \B{L} Q \B{L}\tran \exp\left(F\left(\Delta t - \tau\right)\right)\tran \mathrm{d} \beta
\end{equation}


\begin{equation}
	\mathbb{E}[\B{x}_t \tran \B{x}_t] = \Sigma = \B{L}Q\B{L}\tran - A\B{L}Q\B{L}\tran A\tran
\end{equation}

Which implies that the dynamics for each element of each particle can be updated according to,

\begin{equation}
	\B{x}_{t+1} = A \B{x}_t + \B{d}, \quad \B{d} \sim \mathcal{N}(\B{0}, \Sigma)
\end{equation}

The transition probability from $\B{x}_{t}$ to $\B{x}_{t+1}$ can therefore be written in the form a a Gaussian distribution,

\begin{equation}
	p(\B{x}_{t+1} | \B{x}_{t}) = \mathcal{N}(A \B{x}_t, \Sigma)
\end{equation}

A naïve implementation of the proposed algorithm is sketched in \ref{alg:PAO_bad}. 

\begin{algorithm}
	\caption{Particle attractor optimisation (PAO)}\label{alg:PAO_bad}
	\begin{algorithmic}
		\Require Number of particles $N$, hyperparameters $\{m,\zeta,k_i,q,\Delta t\}$
		\State Initialise an initial swarm of $N$ particle vectors $\B{x}_{j} \in \mathbb{R}^D$
		\State Initialise the initial velocities of the $N$ particle vectors.
		\State Compute the value of the objective function for each particle.
		\For{each generation}
		\For{each particle}
		\State Compute the positions attraction centres $\alpha_r$.
		\State Compute the position of the particle in the transformed coordinates as $\B{x}' = \B{x} - \frac{1}{k'}\sum_r k_r \alpha_{r}$.
		\State Compute the mean and the covariances for the transition density as above.
		\State Sample a new particle from the $\mathcal{N}(A \B{x}_t, \Sigma)$
		\State Recover the positions of the new particle in the original coordinates as $\B{x} = \B{x}' - \frac{1}{k'}\sum_r k_r \alpha_{r}$.
		\State Compute the value of the objective function for the new particle.
		\EndFor
		\EndFor
	\end{algorithmic}
\end{algorithm}

\subsection*{Notes on implementation}

Since the positions of the global and local best particles change in each iteration (and therefore the value of $Q$), it is necessary to recompute $\Sigma$ each time the swarm moves (Note that $A$, and $Q$ only come to depend on fixed hyperparameters.). Unfortunately, the relation,

\begin{equation}
	\Sigma = \B{L}Q\B{L}\tran - A\B{L}Q\B{L}\tran A\tran
\end{equation}

is prone to numerical instability as both terms in the difference are similar in magnitude. Fortunately, the method of \emph{matrix fraction decomposition} can be employed to jointly calculate $A$ and $\Sigma$ employing only a single matrix exponential. The interested reader is directed to \cite{Saerkkae2019} for a full treatment, but the salient result is,

\begin{equation}
	\exp (\Phi \Delta t) = \exp \left(\m{F & \B{L} Q \B{L}\tran \\ \B{0} & -F\tran} \Delta t \right) = \m{A & \Sigma (A^{-1})\tran \\ \B{0} & (A^{-1})\tran}
\end{equation}

Thus, $\Sigma$ can be read off as the upper right block of $\exp (\Phi \Delta t)$ post-multiplied by the upper left block.

Computing the matrix exponential every iteration still adds undesirable computational complexity to the method. In order to significantly alleviate this, one notices that the covariance structure is independent for every element of every particle and can therefore be extracted as a constant in the computation. If one pulls out Q as a factor from the above, the evolution of the dynamics can thus be written,

\begin{equation}
	\B{x}_{t+1} = A \B{x}_t + q_0\nu(\alpha)\B{d}, \quad \B{d} \sim \mathcal{N}(\B{0}, \Sigma)
\end{equation}

where $A$ and $\Sigma$ can be precomputed. Further numerical advantage can be achieved by pre-computing the Cholesky decomposition of $\Sigma$,

\begin{equation}
	\Sigma = HH\tran
\end{equation}

so that the above can be written,

\begin{equation}
	\B{x}_{t+1} = A \B{x}_t + \sqrt{q_0\nu(\alpha)}H\B{d}, \quad \B{d} \sim \mathcal{N}(0,I_2)
\end{equation}

The overall computation (for every element of every particle) can be efficiently implemented on modern architecture using tensor operations. The proposed algorithm is sketched in algorithm \ref{alg:PAO}. In the notation below, $A \odot B$ refers to a broadcasted matrix-vector multiplication between the last two indices of the tensor $A$ and the last axis of the Tensor $B$.  $A \otimes B$ represents a tensor outer product.











\begin{algorithm}
	\caption{An efficient implementation of PAO}\label{alg:PAO}
	\begin{algorithmic}
		\Require Number of particles $N$, hyperparameters $\{m,\zeta,k_i,q,\Delta t\}$
		\State Precompute transition matrix $A \in \mathbb{R}^{2\times 2}$ and covariance $\Sigma \in \mathbb{R}^{2\times 2}$ for a single element of a single particle.
		\State Take the Cholesky decomposition of $\Sigma$ to compute $H \in \mathbb{R}^{2\times 2}$.
		\State Generate $N$ such particle vectors of position $\B{x}_{j} \in \mathbb{R}^D$  and velocity $\dot{\B{x}}_{j} \in \mathbb{R}^D$  according to some initialisation scheme. Assemble these vectors into the tensor $X_{ijk} \in \mathbb{R}^{N\times D\times 2}$.
		\State Compute the value of the objective function for each particle in the tensor $X$.
		\For{each generation}
		\State Compute the $r$ attraction centres $\B{\alpha} \in \mathbb{R}^{r \times N \times D}$ as required.
		\State Transform the positions of the particles by $\B{x}_{ij1}' = \B{x}_{ij1} - \frac{1}{k'}\sum_r k_r \alpha_{rij}$.
		\State Move the swarm by $X'_{t+1} = A \odot X'_t + (q\nu(\B{\alpha})\otimes  H) \odot D, \quad D \in \mathbb{R}^{N\times D \times 2}, \quad d_{ijk} \sim \mathcal{N}(0,1)$.
		\State Recover the positions of the particles in the original coordinates as $\B{x}_{ij1} = \B{x}'_{ij1} + \frac{1}{k'}\sum_r k_r \alpha_{rij}$.
		\State Compute the value of the objective function for each particle in the tensor $X$.
		\EndFor
	\end{algorithmic}
\end{algorithm}


\section{Benchmark comparison study}

In order to demonstrate that the proposed approach is competitive with other approaches a simple benchmark study is presented. Although it is not the primary objective of this algorithm to produce a heuristic method that out-performs state-of-the art methods, it is important to establish that the proposed method remains effective at the task of optimisation and that the benefits of interpretable hyperparameters, exact dynamics and closed form transition densities do not come at the cost of optimisation performance.

In order to assess the performance of PAO, a suite of nine standard optimisation benchmark problems are considered. The problems are selected in order to demonstrate the effectiveness of the model in a number of challenging scenarios. The considered problems are collected in Table \ref{tbl:problems} in approximate order of difficulty.

\begin{table}
	\centering
	\caption{Benchmark optimisation functions considered in this study.}
	\label{tbl:problems}
	\begin{tabular}{llll}
		Problem                  & Domain                    & $f(\B{x})$                                                                                & $\B{x}^*$            \\ \hline
		De Jong's function       & $x_i \in [-5.12, 5.12]$   & $\sum_i^N x_i^2$                                                                          & $x_i = 0$            \\
		Hyper-ellipsoid function & $x_i \in [-5.12, 5.12]$   & $\sum_i^N i x_i^2$                                                                        & $x_i = 0$            \\
		Rotated Hyper-ellipsoid  & $x_i \in [-65.54, 65.54]$ & $\sum_i^n\sum_j^i x_j^2$                                                                  & $x_i = 0$            \\
		Power-sum function       & $x_i \in [-1, 1]$         & $\sum_i^n |x_i|^{i+1}$                                                                    & $x_i = 0$            \\
		Rosenbrock function      & $x_i \in [-2.048, 2.048]$ & $\sum_i^n [100(x_{i+1} - x_i^2)^2 + (1-x_i)^2]$                                           & $x_i = 1$            \\
		Griewangk's  function    & $x_i \in [-600, 600]$     & $\frac{1}{400} \sum_i^n x_i^2 - \prod_i^n \cos(\frac{x_i}{\sqrt{i}}) + 1$                 & $x_i = 0$            \\
		Rastrigin's  function    & $x_i \in [-5.12, 5.12]$   & $10n + \sum_i^n [x_i^2 - 10*\cos(2 \pi x_i)]$                                             & $x_i = 0$            \\
		Ackley's  function       & $x_i \in [-32.77, 32.77]$ & $-20e^{-0.2 \sqrt{\frac{1}{n}\sum_i^n x_i^2}}-e^{\frac{1}{n}\sum_i^n cos(2\pi x_i)}+20+e$ & $x_i = 0$            \\
		Schwefel's  function     & $x_i \in [-500, 500]$     & $\sum_i^n [-x_i \sin(\sqrt{|x_i|})]$                                                      & $x_i \approx 420.97$ \\
	\end{tabular}
\end{table}

For each benchmark function, both two and eight dimensional problems are considered. The performance of the proposed algorithm (PAO) is compared to several very popular optimisation approaches that have proven effective in a wide range of fields:

\begin{itemize}
	\item Particle swarm optimisation (PSO) \cite{Kennedy1995}.
	\item Quantum particle swarm optimisation (QPSO) \cite{Sun2004}.
	\item Differential Evolution (DE) \cite{Storn1997}.
	\item Self-adapting differential evolution (SADE) \cite{Qin2005}.
\end{itemize}

\begin{figure}
	\centering
	\begin{subfigure}[b]{0.33\textwidth}
		\centering
		\includegraphics[width=\textwidth]{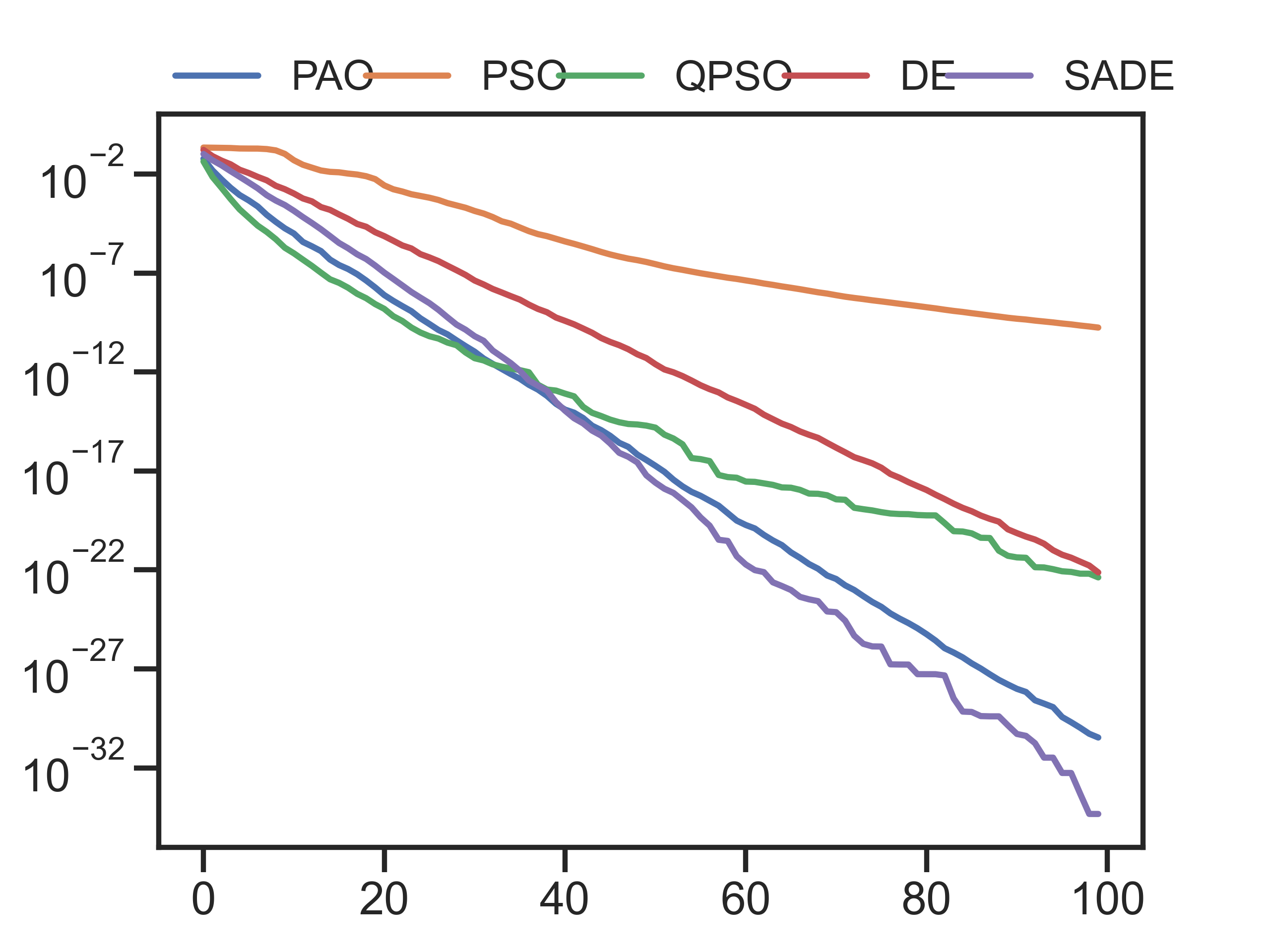}
		\caption{DeJong function.}
		\label{fig:2d_11}
	\end{subfigure}
	\hfill
	\begin{subfigure}[b]{0.33\textwidth}
		\centering
		\includegraphics[width=\textwidth]{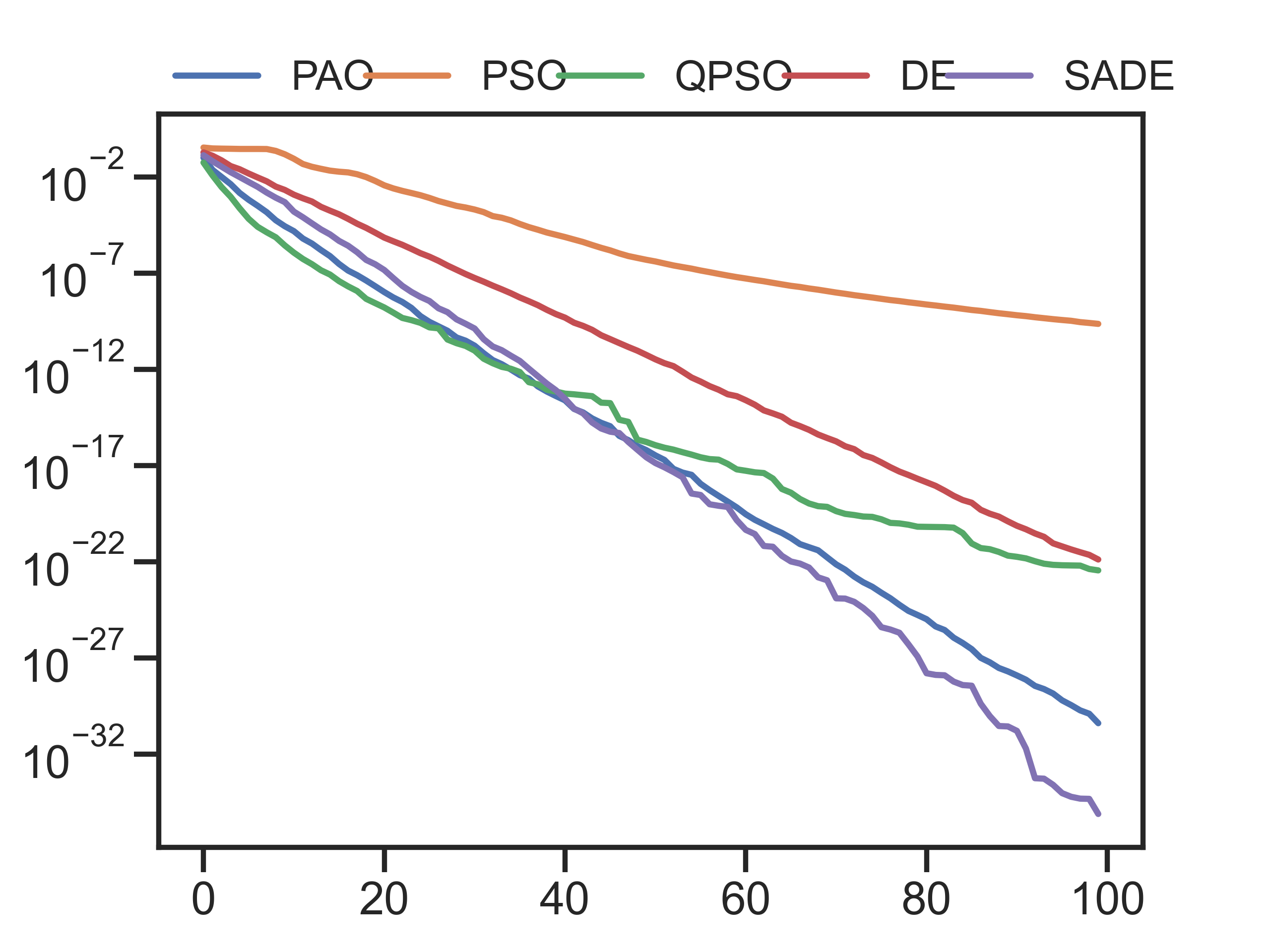}
		\caption{Hyper Ellipsoid function.}
		\label{fig:2d_12}
	\end{subfigure}
	\hfill
	\begin{subfigure}[b]{0.33\textwidth}
		\centering
		\includegraphics[width=\textwidth]{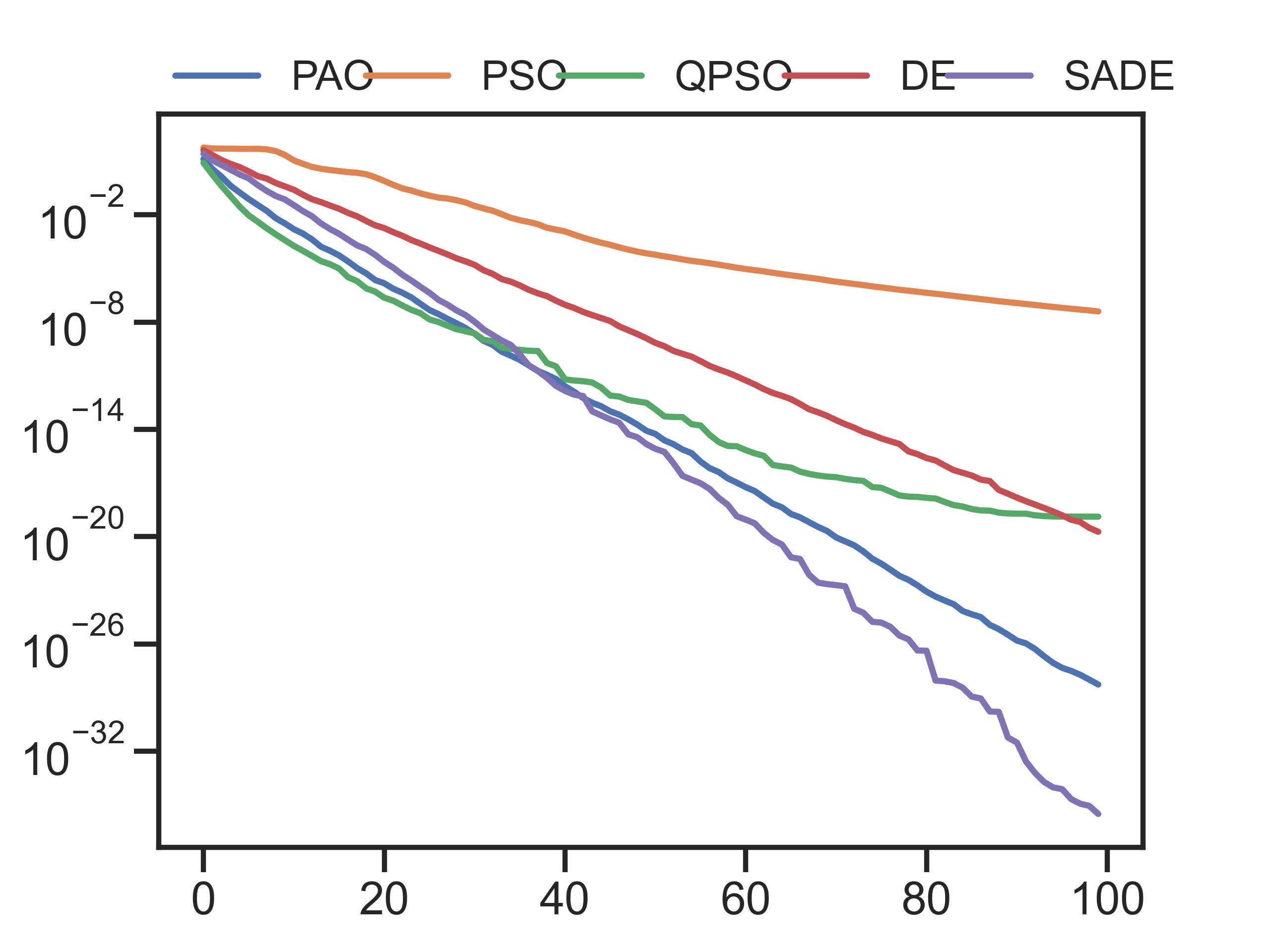}
		\caption{Rotated Hyper Ellipsoid.}
		\label{fig:2d_13}
	\end{subfigure}
	\begin{subfigure}[b]{0.33\textwidth}
		\centering
		\includegraphics[width=\textwidth]{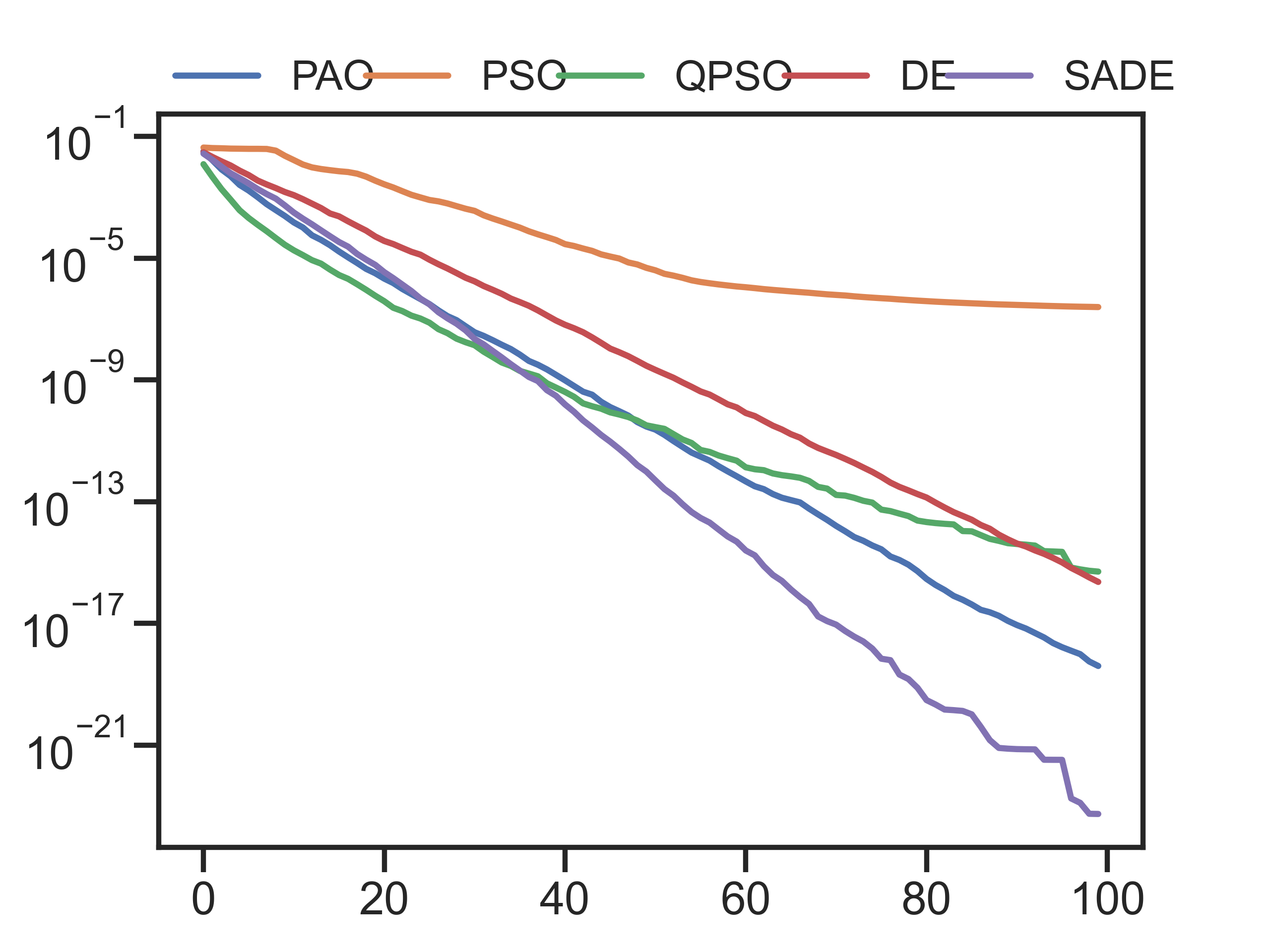}
		\caption{Power sum function.}

		\label{fig:2d_21}
	\end{subfigure}
	\hfill
	\begin{subfigure}[b]{0.33\textwidth}
		\centering
		\includegraphics[width=\textwidth]{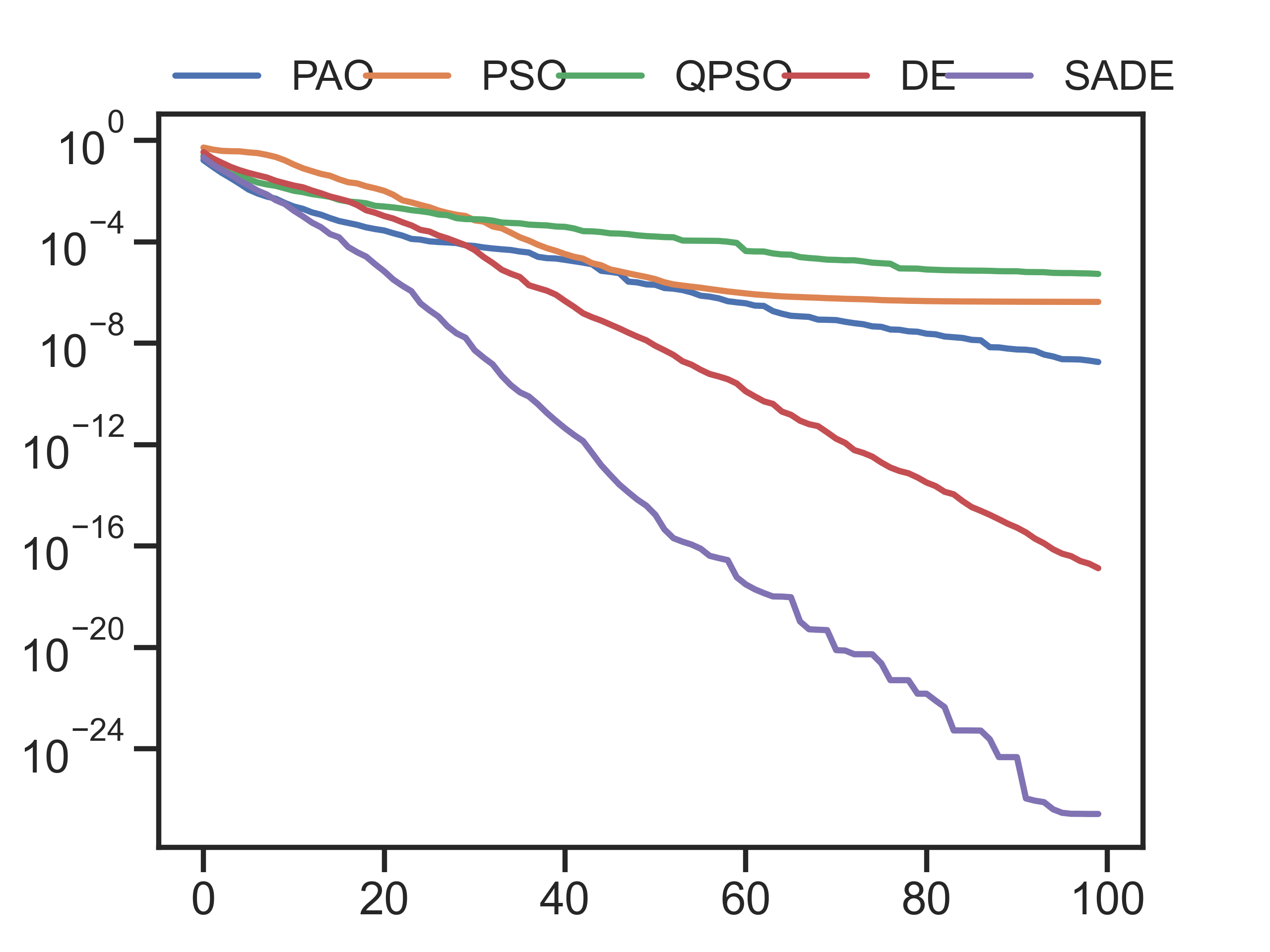}
		\caption{Rosenbrock function.}
		\label{fig:2d_22}
	\end{subfigure}
	\hfill
	\begin{subfigure}[b]{0.33\textwidth}
		\centering
		\includegraphics[width=\textwidth]{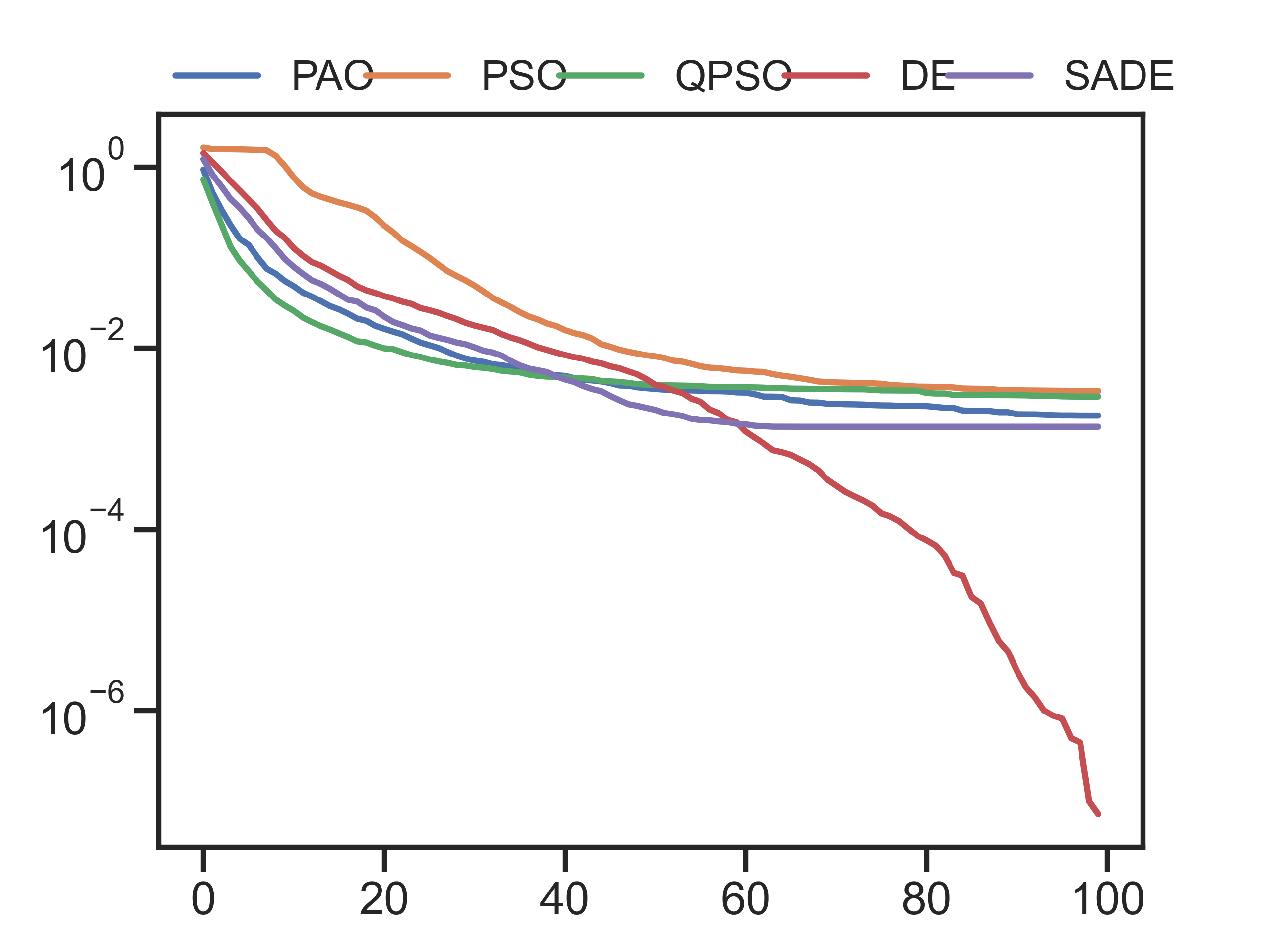}
		\caption{Griewangk function.}
		\label{fig:2d_23}
	\end{subfigure}
	\begin{subfigure}[b]{0.33\textwidth}
		\centering
		\includegraphics[width=\textwidth]{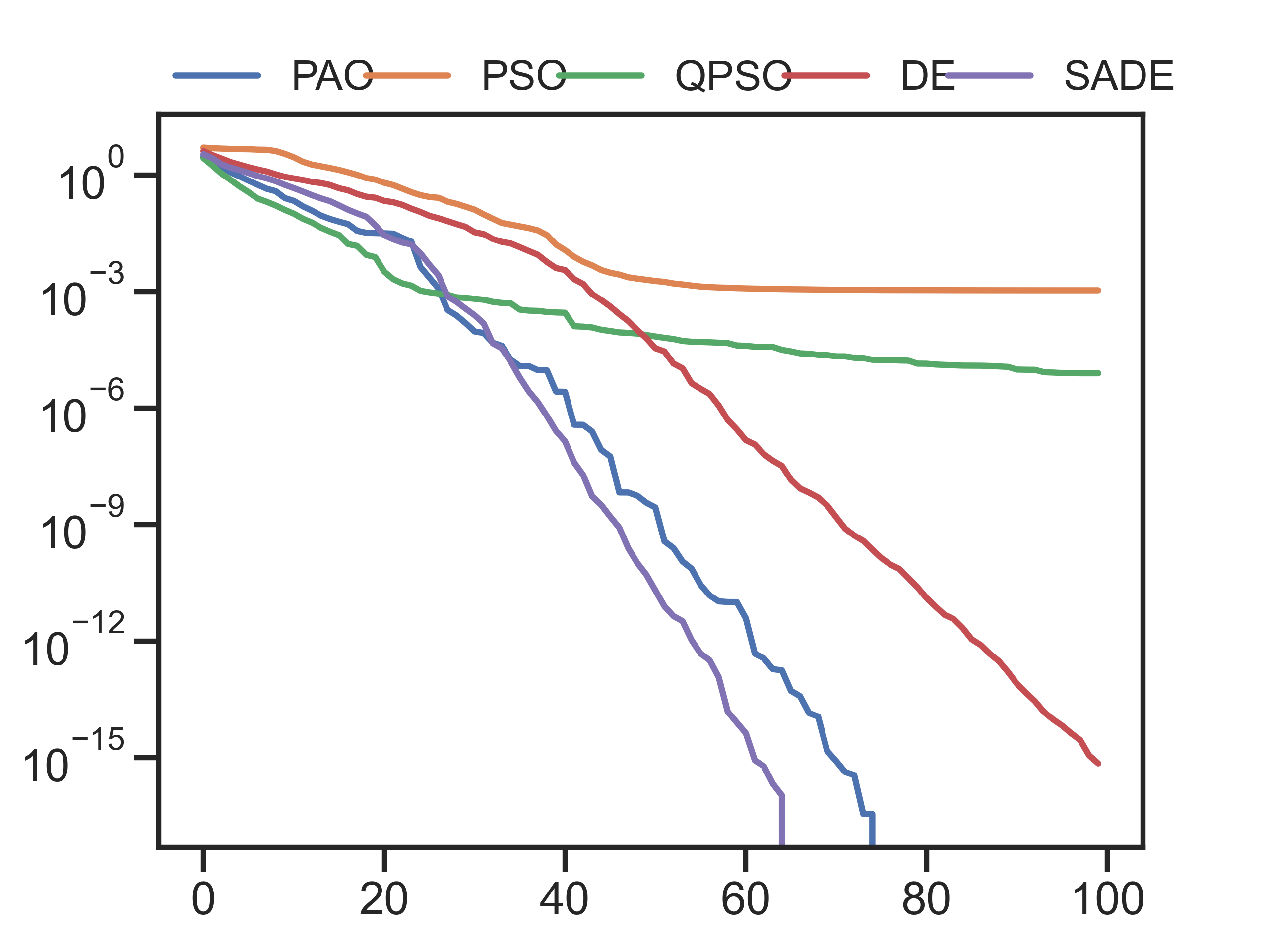}
		\caption{Rastrigin function.}
		\label{fig:2d_31}
	\end{subfigure}
	\hfill
	\begin{subfigure}[b]{0.33\textwidth}
		\centering
		\includegraphics[width=\textwidth]{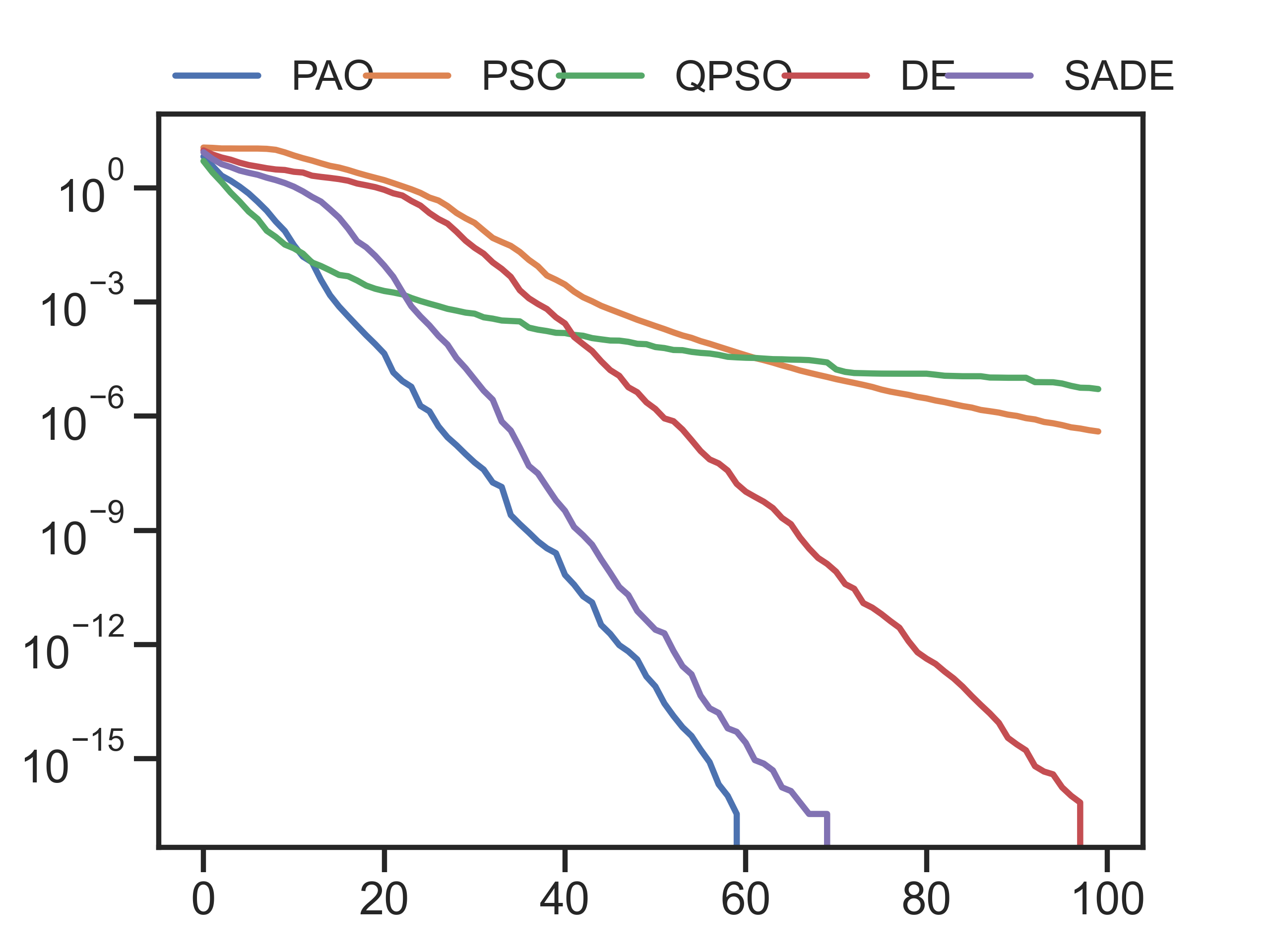}
		\caption{Ackley function.}
		\label{fig:2d_33}
	\end{subfigure}
	\hfill
	\begin{subfigure}[b]{0.33\textwidth}
		\centering
		\includegraphics[width=\textwidth]{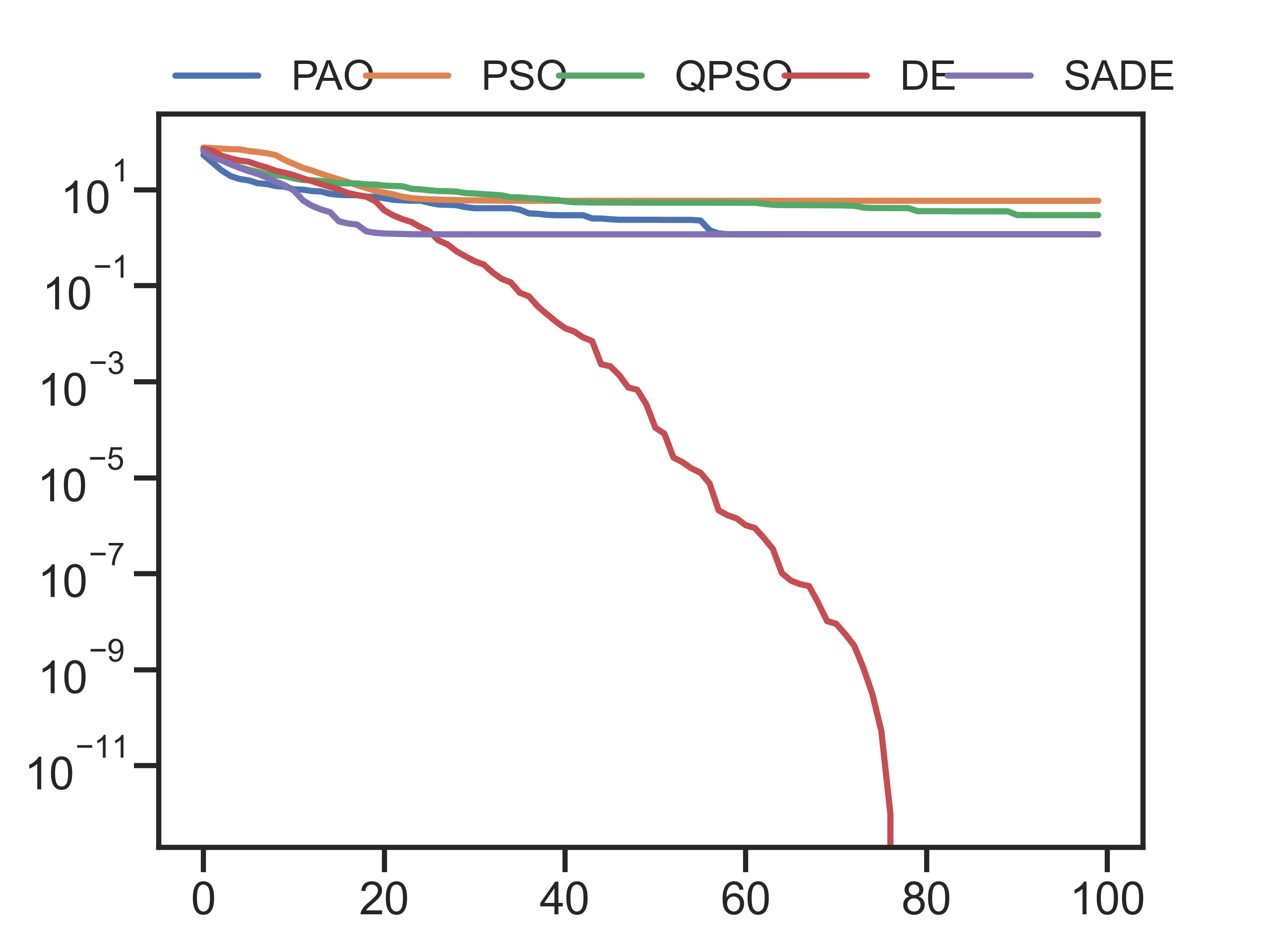}
		\caption{Schwefel function.}
		\label{fig:2d_32}
	\end{subfigure}
	\caption{Convergence histories of the best particle (averaged over 100 runs of each optimiser) for the Molka benchmark suite in 2D.}
	\label{fig:Molka_2D}
\end{figure}

\begin{figure}
	\centering
	\begin{subfigure}[b]{0.33\textwidth}
		\centering
		\includegraphics[width=\textwidth]{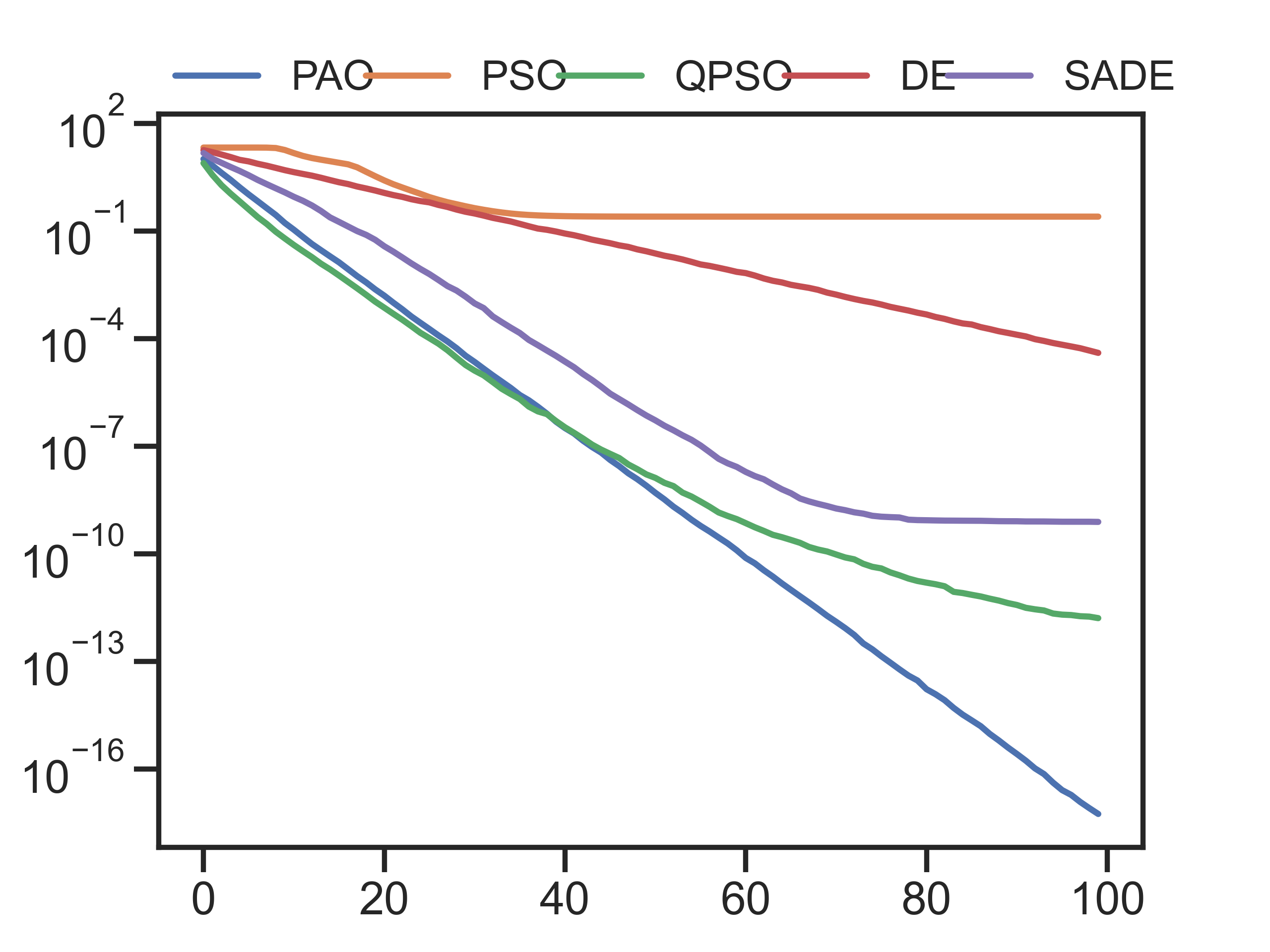}
		\caption{DeJong function.}
		\label{fig:8d_11}
	\end{subfigure}
	\hfill
	\begin{subfigure}[b]{0.33\textwidth}
		\centering
		\includegraphics[width=\textwidth]{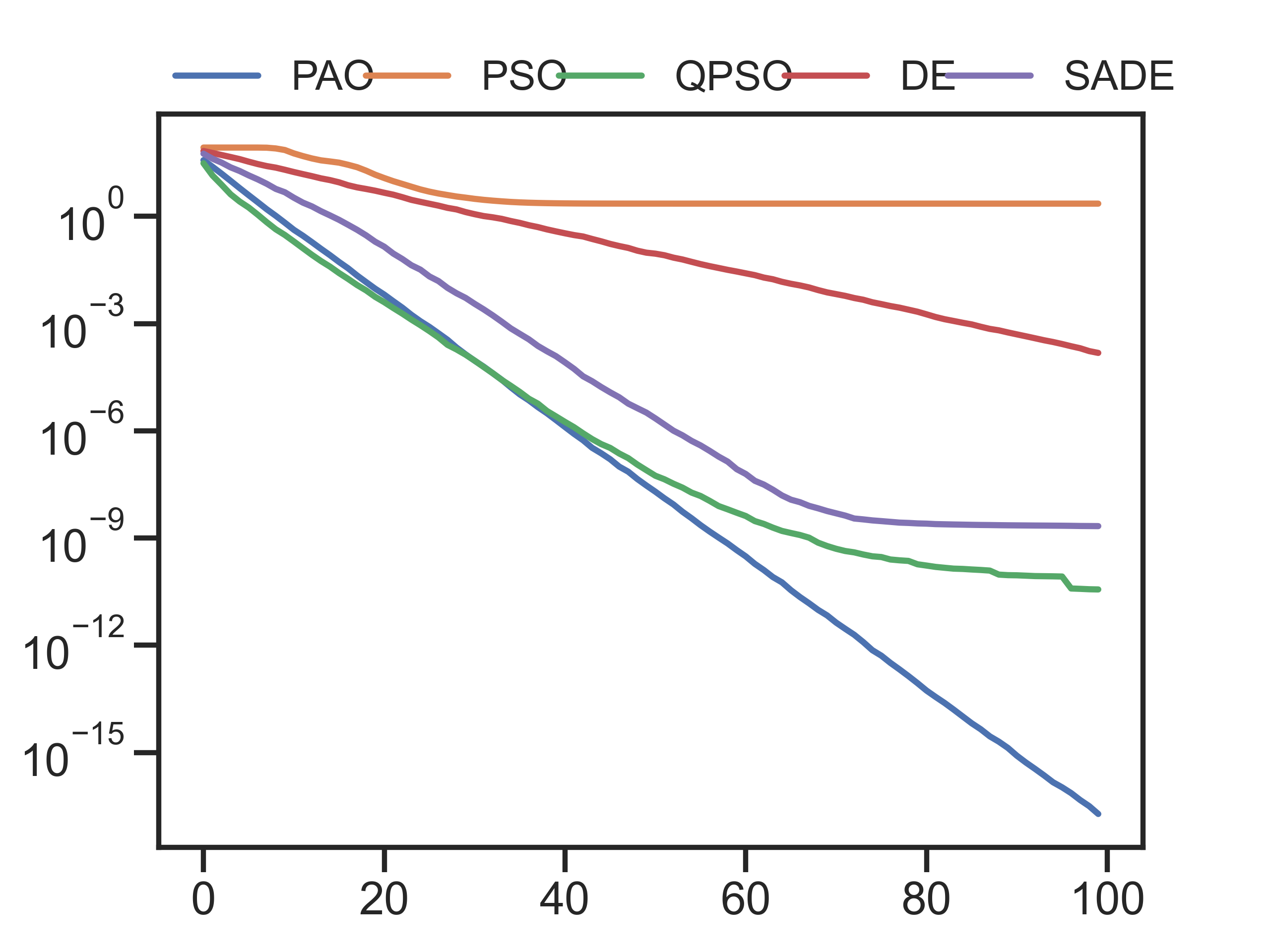}
		\caption{Hyper Ellipsoid function.}
		\label{fig:8d_12}
	\end{subfigure}
	\hfill
	\begin{subfigure}[b]{0.33\textwidth}
		\centering
		\includegraphics[width=\textwidth]{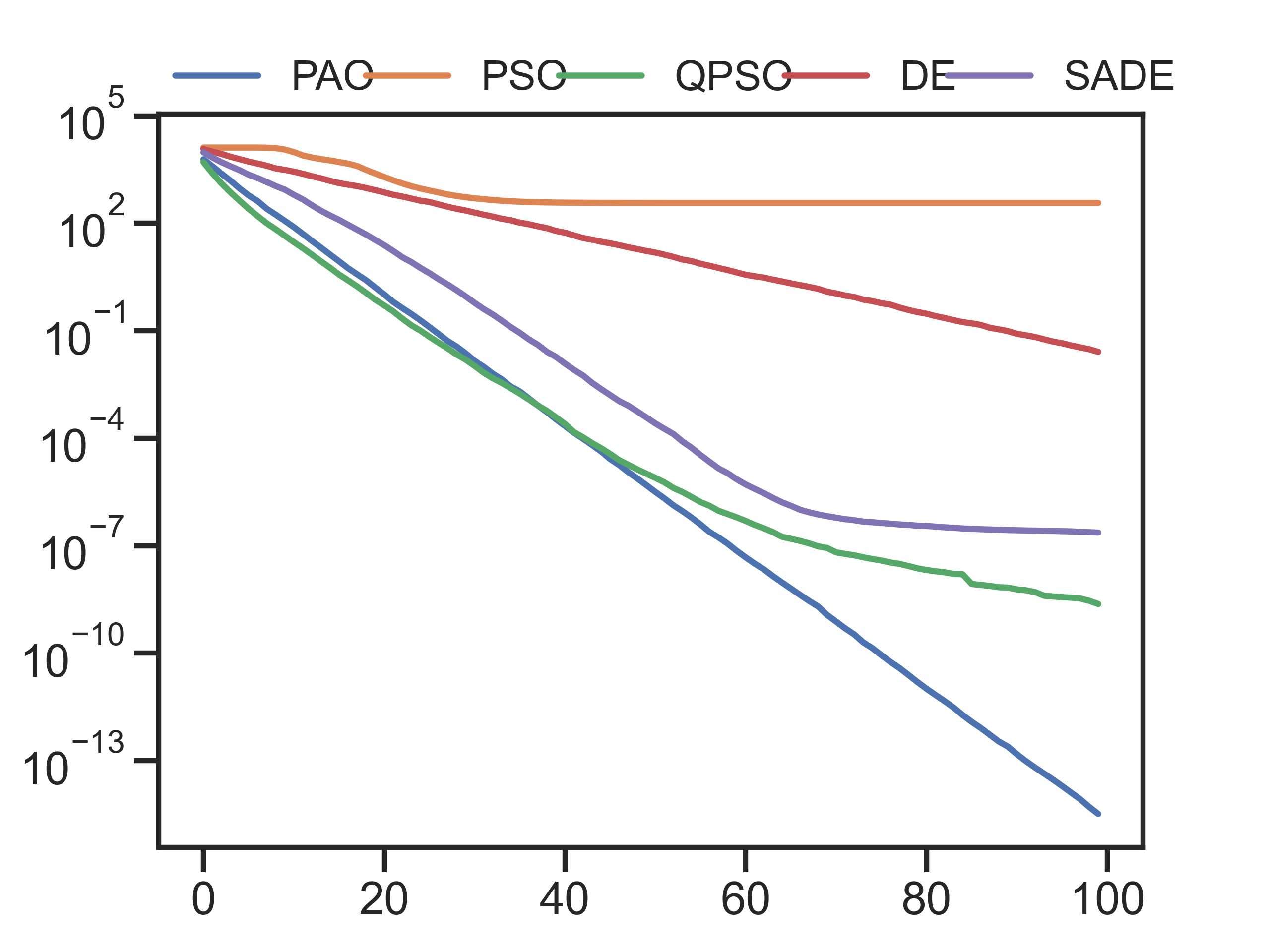}
		\caption{Rotated Hyper Ellipsoid.}
		\label{fig:8d_13}
	\end{subfigure}
	\begin{subfigure}[b]{0.33\textwidth}
		\centering
		\includegraphics[width=\textwidth]{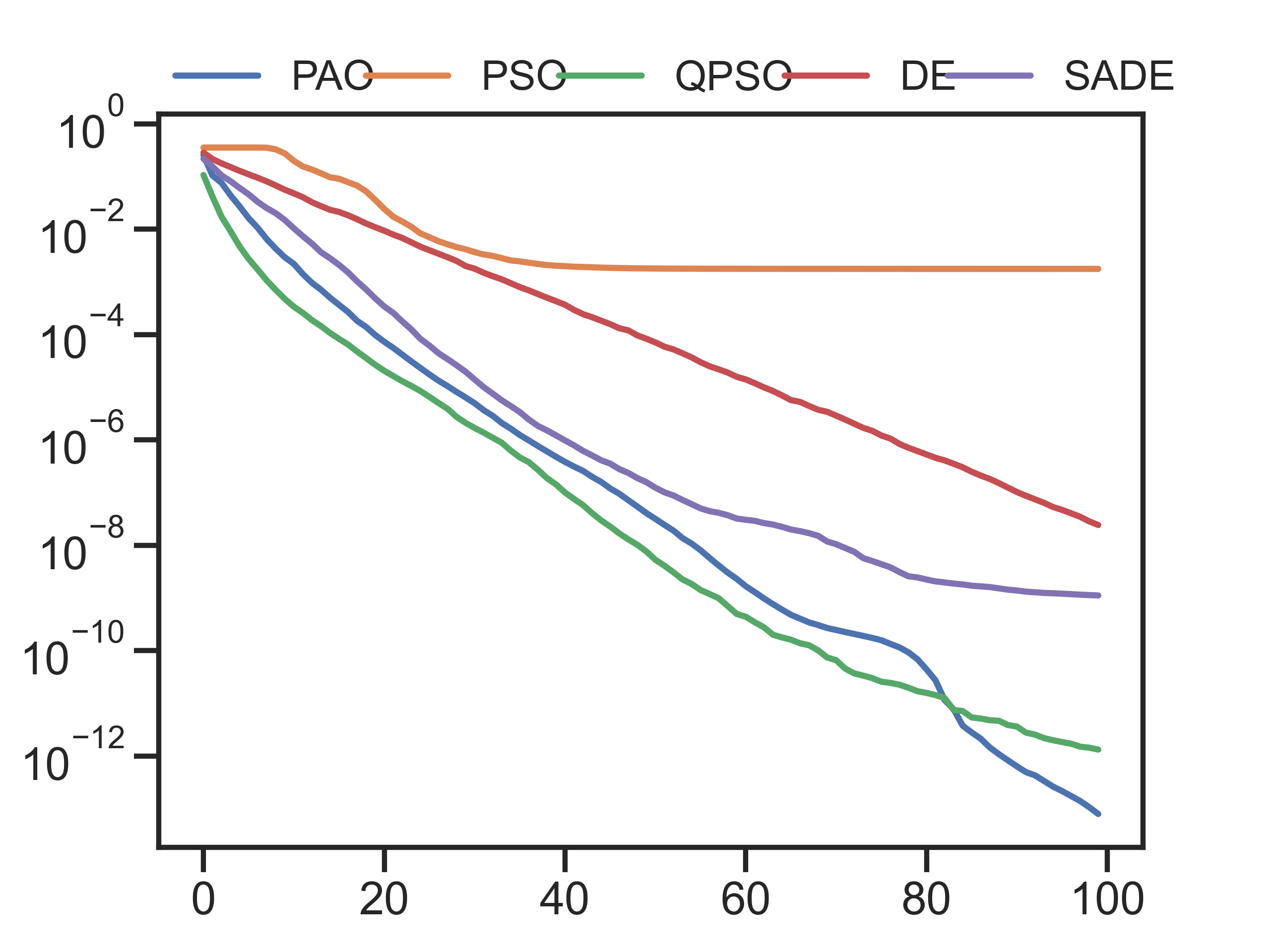}
		\caption{Power sum function.}
		\label{fig:8d_21}
	\end{subfigure}
	\hfill
	\begin{subfigure}[b]{0.33\textwidth}
		\centering
		\includegraphics[width=\textwidth]{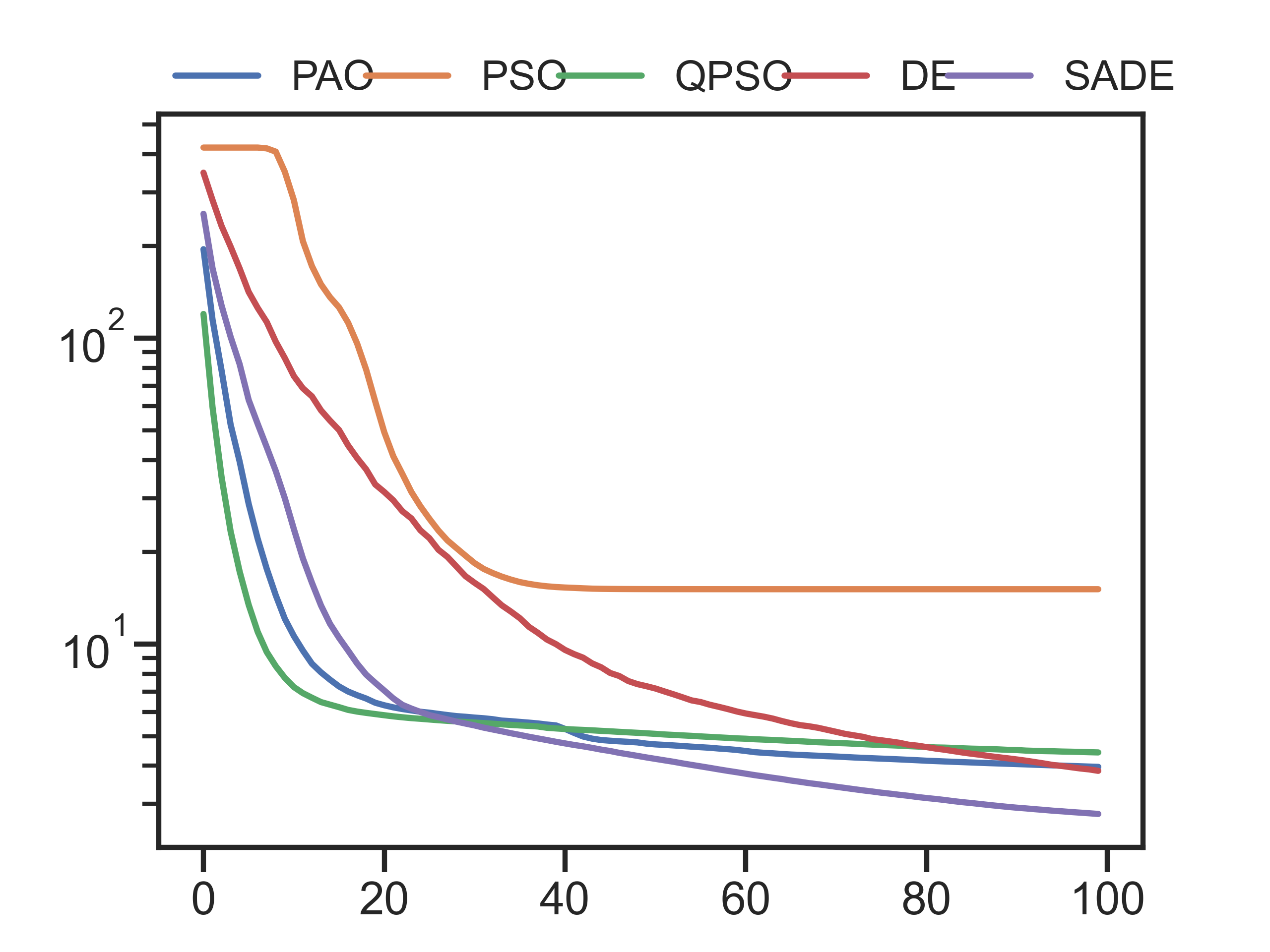}
		\caption{Rosenbrock function.}
		\label{fig:8d_22}
	\end{subfigure}
	\hfill
	\begin{subfigure}[b]{0.33\textwidth}
		\centering
		\includegraphics[width=\textwidth]{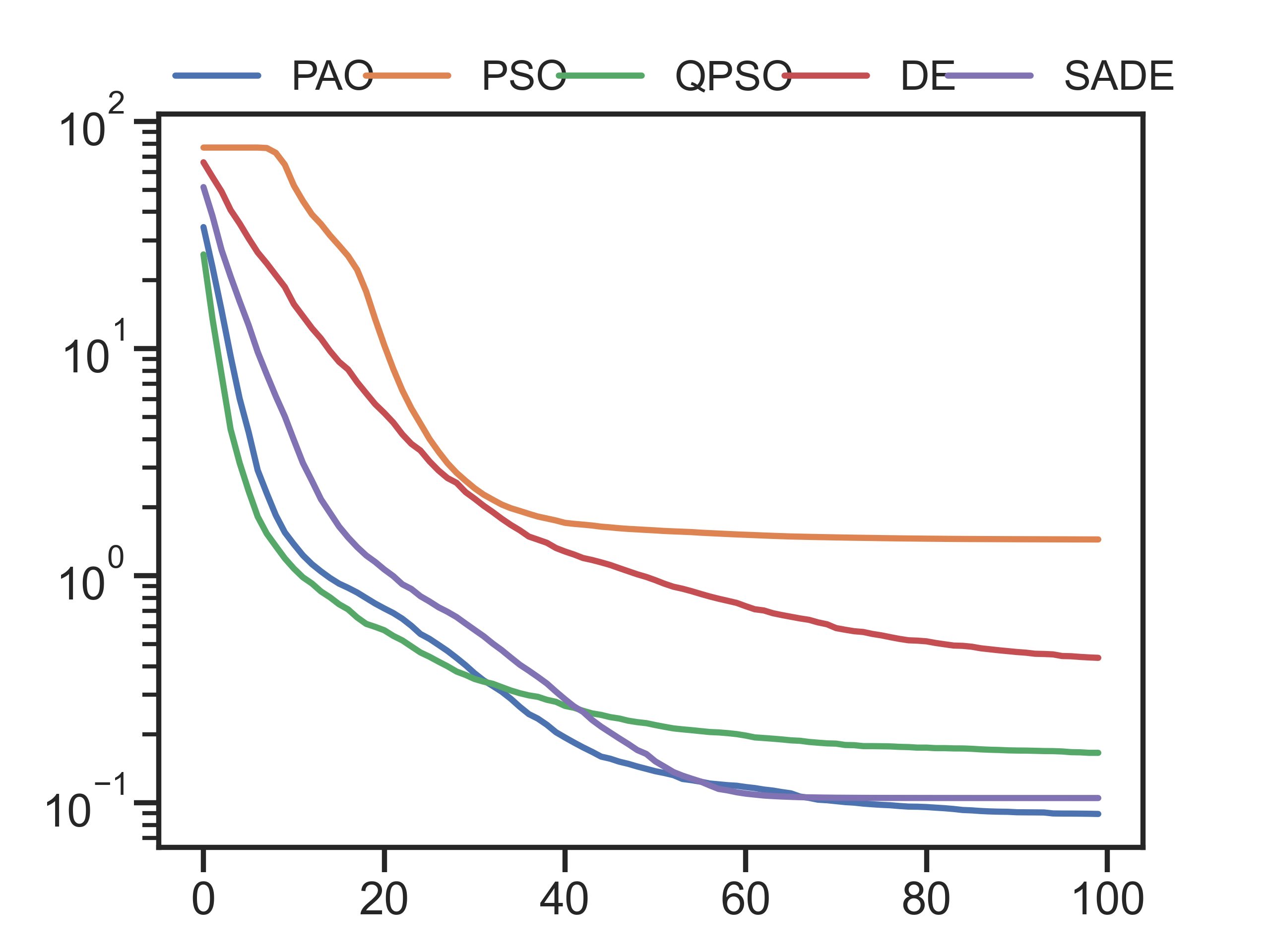}
		\caption{Griewangk function.}
		\label{fig:8d_23}
	\end{subfigure}
	\begin{subfigure}[b]{0.33\textwidth}
		\centering
		\includegraphics[width=\textwidth]{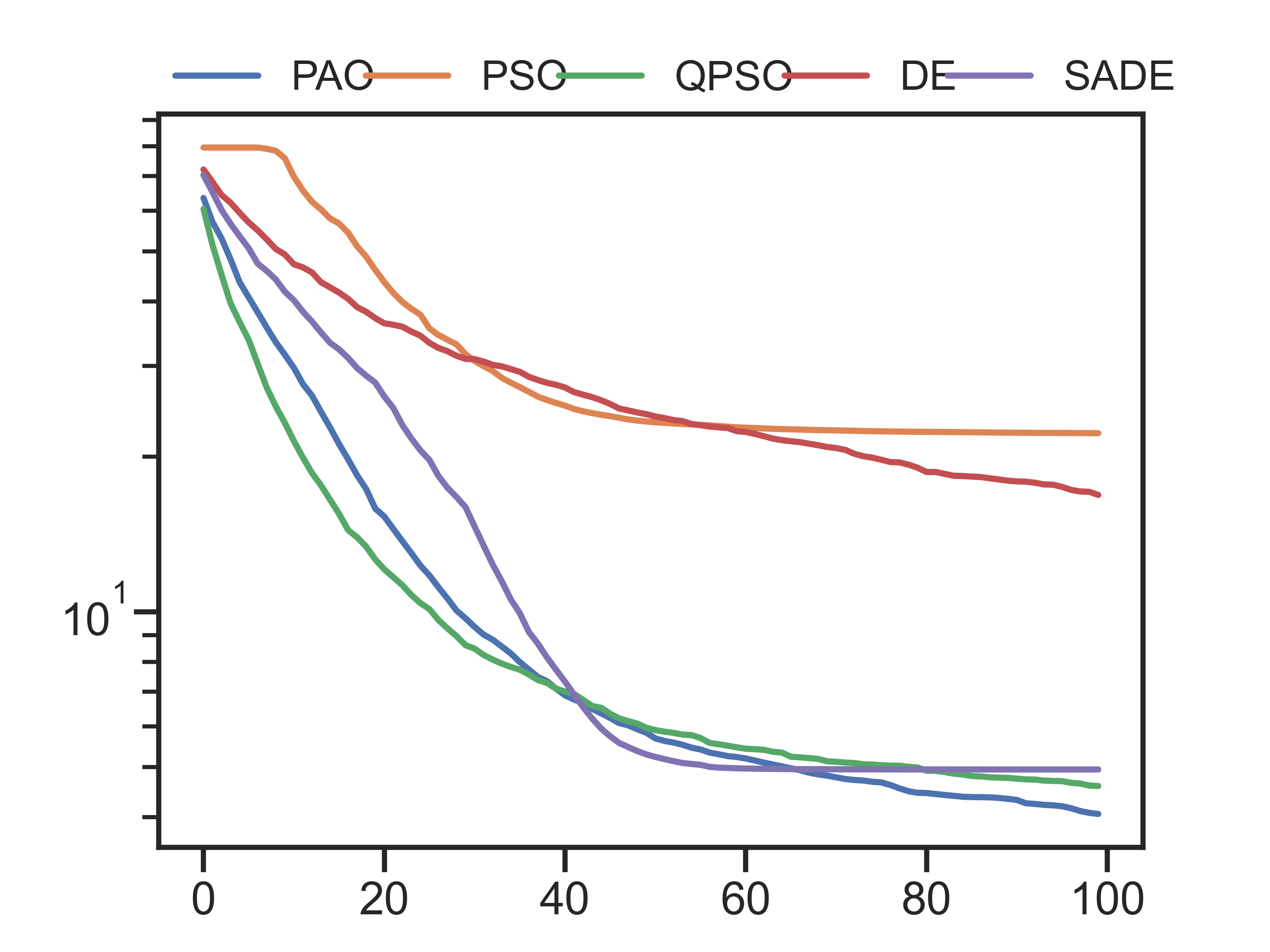}
		\caption{Rastrigin function.}
		\label{fig:8d_31}
	\end{subfigure}
	\hfill
	\begin{subfigure}[b]{0.33\textwidth}
		\centering
		\includegraphics[width=\textwidth]{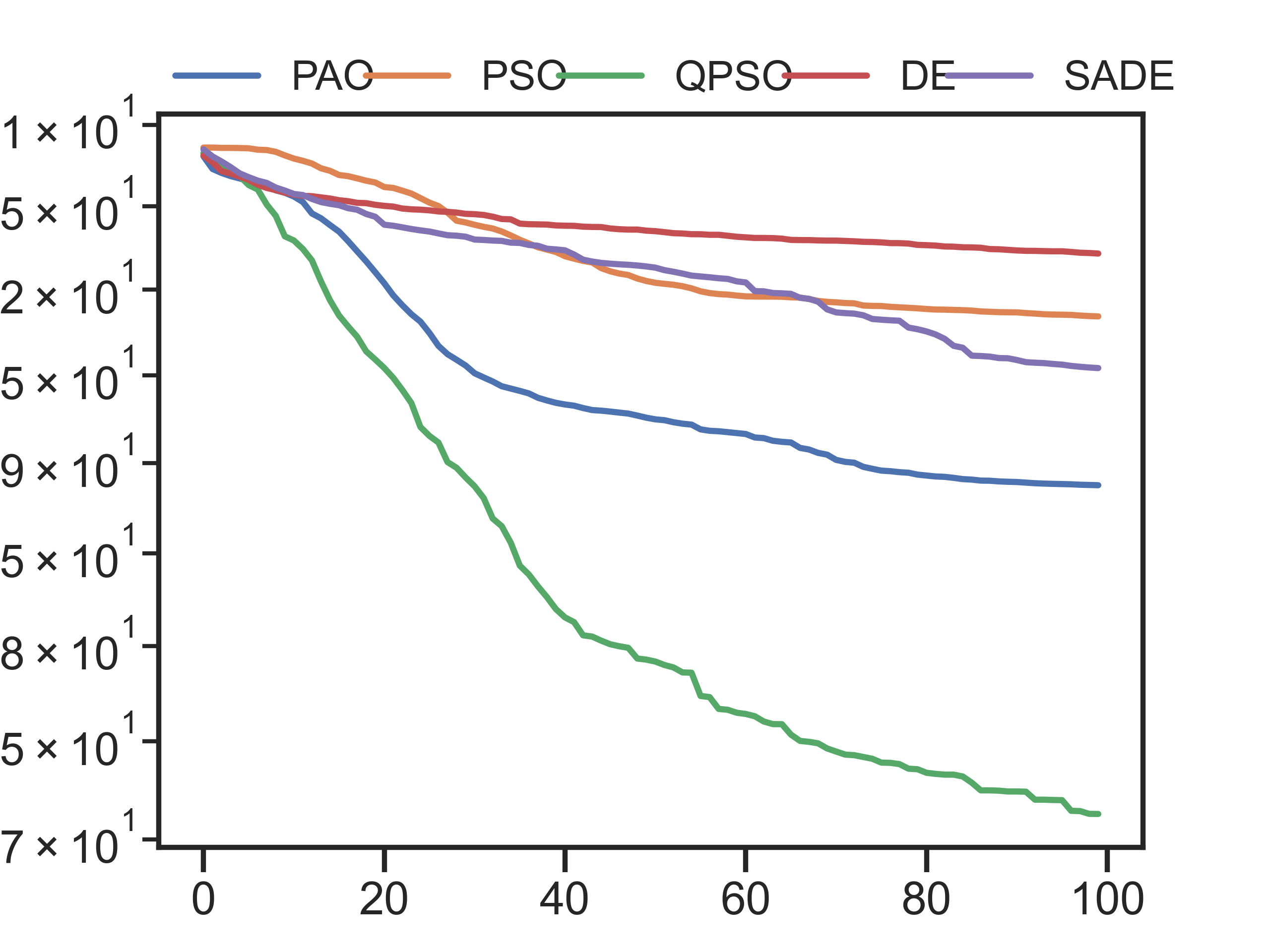}
		\caption{Ackley function.}
		\label{fig:8d_33}
	\end{subfigure}
	\hfill
	\begin{subfigure}[b]{0.33\textwidth}
		\centering
		\includegraphics[width=\textwidth]{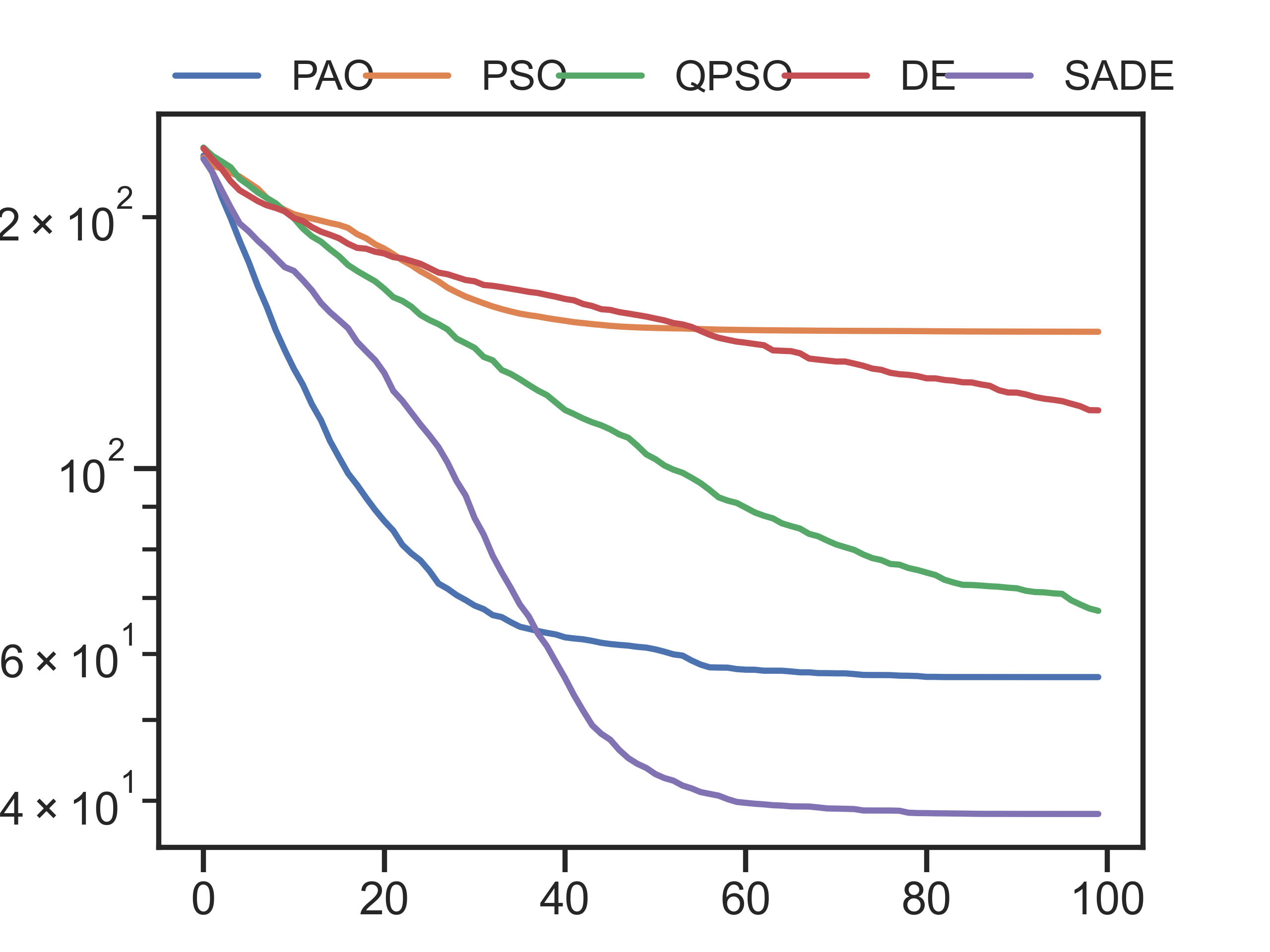}
		\caption{Schwefel function.}
		\label{fig:8d_32}
	\end{subfigure}
	\caption{Convergence histories of the best particle (averaged over 100 runs of each optimiser) for the Molka benchmark suite in 8D.}
	\label{fig:Molka_8D}
\end{figure}

All of the above algorithms (including PAO) are implemented with default hyperparameters from the freelunch python package developed by the authors.\footnote{Implementations of all algorithms used in this study are freely available as part of the open-source optimisation library `FreeLunch', available at \url{https://pypi.org/project/freelunch/}.} To best highlight the performance of the PAO approach, the attractors used in this study are simply the local best and global best particle locations as in the standard PSO algorithm. For each of the nine benchmark functions, each optimiser was randomly initialised 100 times and run for 100 generations with a population size of 100. In all cases this corresponds to 100 runs of each optimiser with a budget of $1.01\times10^3$ function evaluations. Figures \ref{fig:Molka_2D} and \ref{fig:Molka_8D} depict convergence histories averaged over the 100 runs for each algorithm on each optimisation problem for the 2D and 8D suites respectively. To aid comparison, the convergence histories in the figures are shifted such that value of the objective function at the true optimum lies at zero, i.e. $f(\B{x}^*) = 0$. For the convenience of the reader, all parameters relating to the benchmark study and PAO are collected in Table \ref{tbl:params}

\begin{table}
	\centering
	\caption{Parameters relating to the benchmark comparison study.}
	\label{tbl:params}
	\begin{tabular}{lll}
		Parameter  & Description                         & Value \\ \hline
		$m$        & PAO inertia coefficient             & 1.0   \\
		$\zeta$    & PAO damping ratio                   & 0.2   \\
		$k_r$      & PAO stiffness parameters            & 1.0   \\
		$q_0$      & PAO stochastic parameter            & 1.0   \\
		$\Delta t$ & PAO integration interval            & 1.0   \\
		$N$        & Population size (all optimisers)    & 100   \\
		$G$        & Iterations per run (all optimisers) & 100   \\
	\end{tabular}
\end{table}

\section{Discussion}

As can be seen from the figures, the PAO algorithm has performed well on the benchmark optimisation problems. In the 2D set of problems, the performance of PAO on the continuous, unimodal problems (from DeJong to Power sum inclusive) is excellent, surpassed only by the SADE method which is more computationally demanding. On the nonconvex and multimodal problems, the performance of PAO is very much in line with the other approaches, finding exact optima (within numerical tolerance) on both the Rastrigin and Ackley problems.

In the 8D suite of problems, the performance of PAO is excellent on the continuous multimodal benchmarks, clearly outstripping the other methods. In the other problems PAO performs similarly to other methods and manages to avoid becoming trapped in poor minima on the Rastrigin and Griewangk functions as is observed with the PSO and DE methods. In the challenging Schwefel and Ackley functions, almost all methods (PAO included) were unable to locate the global minima in the higher dimensional problems with default hyperparameters.

The benchmark results are especially impressive given that the attractors in PAO implementation were simply set to the be local and global best particles as in the standard PSO algorithm. Indeed, PAO has outperformed the PSO approach in every problem. It might be expected that the the selection of more specialised attractors may enable the practitioner to input domain knowledge into the optimisation procedure resulting in greater performance. For example, one could imagine that local minima might be avoided by using a stochastic attractor (in which the position of the attractor is drawn from some distribution for each particle) when the problem is known to be highly multi-modal.

Although the benchmark performance is encouraging, the authors would make the argument that it is not the only feature that should be considered in the assessment of a novel heuristic. As well as optimisation power, PAO has been shown to have a number of desirable features including interpretable hyperparameters, exact dynamics and closed-form transition densities.

The hyperparameters of the PAO approach are familiar to engineers as the inertial, damping and stiffness terms of a linear dynamic system. This choice is deliberate on the part of the authors in order to permit an interpretable way to select hyperparameters. Large choices for the inertial parameter slow converge and encourage exploration. Whereas large choices for the values of the stiffness parameters accelerate convergence and promote exploitation. Particularly interesting is the choice of the damping ratio parameter $\zeta$. As is the case for linear dynamics, the selection of values  $\zeta<1$ give rise to underdamped dynamics and oscillation, values $\zeta>1$ lead to overdamped dynamics and slow convergence. The choice of the stochastic scaling parameter $q$ can be related to the excitation level of the dynamics and controls the extent to which the PAO algorithm is dominated by random search.

An advantage of PAO compared to other PSO methods is that the motions of the particles can be expressed exactly in terms of a time-step $\Delta t$. Unlike methods that rely on discretisation to move the particles in successive iterations, the exact nature of the dynamics means that no error is accrued even as the value of $\Delta t$ becomes very large. This means that the $\Delta t$ could conceivably be altered during the run in either a scheduled or an adaptive fashion similar to the energy function in simulated annealing \cite{Kirkpatrick1983}. A scheduling approach could used to promote large movements during the initial stages of the optimisation procedure (thus quickly moving to promising regions of the search space) and then more fine-grained motions close to convergence. In highly multi-modal problems, the opposite approach might be taken in order to attempt to escape minima as the particles converge.

A major advantage of the proposed approach is that the transition densities (parametrised by the integration interval $\Delta t$) are available in closed form. One particular advantage is that at the end of the optimisation run, the final transition density (averaged over every particle, weighted by objective score) might be viewed as a kind of uncertainty on the position of the optima. This could provide the practitioner with insight in the case of noisy objective functions such as those that are typically encountered when performing parameter identification tasks from data.

A more formal treatment of uncertainty can also be envisaged. Access to the transitions densities would allow the PAO method to be sued as the proposal step within an SMC scheme. This would enable a Bayesian viewpoint to be taken and would give access to valid posterior distributions over the positions of optima. The application of PAO within an SMC framework has the potential to drastically improve the convergence of SMC algorithms in line with other heuristic proposal schemes such as the DE-based approach in \cite{Turner2012}.

In this paper, a novel variant of the particle swarm algorithm has been proposed with a number of distinct advantages including interpretable hyperparameters, exact dynamics and closed-form access to the transition density. The performance of the proposed approach has been shown to be in line with other common choices for heuristic optimisation, indicating the advantages of the proposed approach do not come at the cost of reduced performance.

\section*{Author contributions}

The authors contributed equally to the conception and development of the method. MDC collected the data and wrote the manuscript.

\section*{Acknowledgements}

The authors would like to thank the Isaac Newton Institute for Mathematical Sciences, Cambridge, for support and hospitality during the programme Data Driven Engineering (DDE) where work on this paper was undertaken. This work was supported by EPSRC grant no EP/W002140/1.

\bibliographystyle{unsrtnat_keith}
\bibliography{PAO.bib}

\end{document}